\documentclass[conference]{IEEEtran}

\usepackage[T1]{fontenc}
\usepackage[utf8]{inputenc}
\usepackage[numbers,sort&compress]{natbib}
\usepackage{amsmath,amssymb,amsfonts}
\usepackage{graphicx}
\usepackage{booktabs}
\usepackage{array}
\usepackage{multirow}
\usepackage{url}
\usepackage{microtype}
\usepackage{hyperref}
\hypersetup{
    colorlinks   = true,
    citecolor    = blue,
    urlcolor    =  blue
}
\usepackage{placeins}

\newcolumntype{L}[1]{>{\raggedright\arraybackslash}p{#1}}
\usepackage{xcolor}
\usepackage{tikz}
\usetikzlibrary{matrix,positioning}

\newcommand{\hf}[2]{\raisebox{-2.2pt}{\includegraphics[scale=0.09]{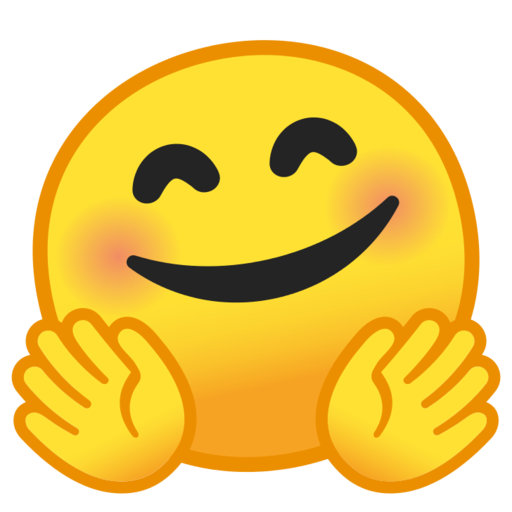}}~\href{#1}{\texttt{#2}}}

\newcommand{\gh}[2]{\raisebox{-2.2pt}{\includegraphics[scale=0.02]{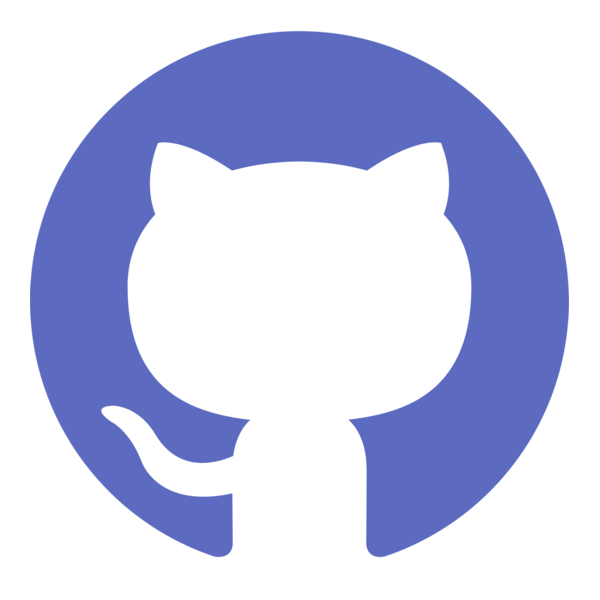}}~\href{#1}{\texttt{#2}}}

\begin{document}

\title{Fault of Our Stars: Behavioral Drivers of Rating–Sentiment Incongruence}

 \author{
 \IEEEauthorblockN{
Ramanaish Abaiyan, Ruththiragayan Sutharsan, Kusal Amantha, Anusan Krishnathas, Asma Rauff,\\
Kovindarajah Sriyathurshan, Patalee Narasinghe, Nirasha Munasinghe, Nisansa de Silva, Sandareka Wickramanayake 
 }
 \IEEEauthorblockA{\textit{Department of Computer Science and Engineering}, \textit{University of Moratuwa}, Moratuwa, Sri Lanka \\
 \{abaiyanr.23, ruththiragayans.23, kusala.23, anusank.23, asmar.23,\\ sriyathurshank.23, patalee.21, nirasha.25, NisansaDdS, sandarekaw\}@cse.mrt.ac.lk}
 }

\maketitle

\begin{abstract}
When people share experiences online, they often express thoughts in two ways: a star rating and a written review. In sentiment analysis, ratings are widely used as convenient weak labels for textual sentiment, yet whether the two actually agree is rarely questioned. This study investigates sentiment–rating incongruence, where the sentiment expressed in review text differs from the sentiment implied by the assigned star rating, in Sri Lankan tourism attraction reviews. A dataset of 16,156 reviews from 2010 to 2023 is analyzed using a transformer-based sentiment pipeline that derives textual sentiment independently of assigned ratings. Incongruence occurs in 18.6\% of reviews and falls into six directional patterns, with Conservative Rater and Obligatory 5-Star behaviors accounting for the majority of mismatches. Prevalence also varies across venue types, with museums showing the highest rates. Statistical tests, logistic regression, Random Forest, and SHAP analysis identify venue type, reviewer expertise, review length, and temporal factors as contributors to rating–text divergence. Overall, this study demonstrates that star ratings are not interchangeable with textual sentiment and should be validated before being treated as ground-truth labels in NLP.
\end{abstract}


\section{Introduction}
Online tourism reviews are an important source of user-generated content for understanding visitor experiences. Most review platforms allow users to express their experience through both a star rating and a written review. In sentiment analysis and review mining, star ratings are often treated as convenient weak labels for textual sentiment \cite{Pang,alaei}. However, this assumption is not always reliable. A high rating does not necessarily mean that the review text is fully positive, and a moderate rating may still contain strongly positive language. For instance, a review reading "Beautiful gardens, but overpriced and overcrowded" accompanied by a 5-star rating illustrates this tension: the text conveys mixed-to-negative sentiment, while the numerical score signals unambiguous satisfaction. This creates an NLP problem: ratings may introduce noisy or context-biased labels when used as ground truth for sentiment analysis.

The growth of platforms such as TripAdvisor has expanded tourism review data \cite{George}. Many studies use these reviews to analyze destination image, tourist satisfaction, and consumer behavior. However, the relationship between rating-derived and text-derived sentiment remains insufficiently examined. In many sentiment analysis pipelines, star ratings are used as proxy labels without validating whether the written review expresses the same polarity. From an NLP perspective, this becomes a weak-supervision problem, where models trained or evaluated using ratings may learn distorted patterns rather than the actual sentiment expressed in language.

Previous studies suggest that rating--text inconsistency is a recurring issue in online reviews \cite{Bigne,Kwon}. However, much of the tourism sentiment analysis literature still relies on ratings as sentiment labels, particularly in hotel and restaurant contexts \cite{alaei,Ameur}. Although aspect-based sentiment analysis and topic modeling have improved the extraction of fine-grained information from review text \cite{Chu2022,Ali}, the broader question of whether ratings reliably represent textual sentiment has received less attention. This gap is particularly important in underrepresented tourism contexts, where reviews are shaped by culture, attraction types, and reviewer experience. 

Recent transformer-based language models provide an opportunity to study sentiment--rating incongruence more effectively. Models such as BERT \cite{Devlin,Sun} and RoBERTa capture contextual meaning more effectively than traditional lexicon-based approaches \cite{Wen,Puh}. In this study, transformer-based sentiment inference is used to derive textual sentiment independently from assigned ratings, allowing review text to be analyzed as a separate linguistic signal.

Using 16,156 Sri Lankan tourism attraction reviews collected between 2010 and 2023 \cite{Sewwandi}, this study investigates how often rating-derived sentiment and NLP-derived textual sentiment diverge, what directional forms these mismatches take, and which contextual and reviewer-level factors are associated with them. Star ratings are grouped into negative, neutral, and positive classes, while textual sentiment is inferred using a transformer-based sentiment pipeline selected through comparative model evaluation. The resulting mismatches are organized into six directional incongruence patterns, moving beyond a simple matched/mismatched classification.

After deriving textual sentiment through transformer-based NLP inference, statistical and machine learning models are used as secondary explanatory tools to examine factors associated with rating--text divergence.This study makes four contributions. First, unlike prior work that treats rating–text mismatch as a binary correction problem, it introduces six directional incongruence patterns that capture how sentiment and ratings diverge, not just whether they do. Second, while transformer-based sentiment inference has been applied to tourism reviews before, this study is among the first to apply it to Sri Lankan tourism attraction data, a setting with limited prior coverage compared to hotel- and restaurant-focused studies. Third, rather than focusing on a single venue type or platform, this study examines incongruence across 11 attraction types over a 13-year span, identifying venue type, reviewer expertise, and temporal trends as structural correlates of mismatch. Finally, it reframes rating–text divergence as a weak-label reliability problem in NLP, rather than primarily a tourism-analytics or review-correction proble, directly questioning the common practice of using ratings as ground-truth sentiment labels in downstream NLP pipelines.

The findings indicate that sentiment–rating incongruence is a meaningful, context-dependent signal and that star ratings should not be treated as ground-truth sentiment without validation. For tourism review analytics, they show that ratings and written reviews capture different aspects of visitor experience, supporting the need for context-aware sentiment analysis approaches.
\hf{https://huggingface.co/datasets/Abaiyan/Sri-lankan-tourism-review-incongruence}{Data} and \gh{https://github.com/Abaiyan-27/Group-J---Research-Paper.git}{code} are publicly available.

\section{Related Work}
The foundational survey by~\citet{Pang} established sentiment analysis as a
major research area and reinforced the common assumption that star ratings
broadly reflect the sentiment expressed in review text. Despite recognized
limitations, this convention remains widely used in tourism research as a
practical weak-labeling strategy. \citet{alaei} note that ratings are
often treated as weak labels without explicit validation, and that the
literature has been heavily concentrated on hotels and restaurants. This
imbalance is further highlighted by~\citet{Ameur}, who report limited venue
diversity and restricted geographic coverage in existing studies, especially
for emerging tourism destinations.

Methodological advances have substantially improved sentiment analysis in
tourism. \citet{Wen} demonstrate the effectiveness of transformer-based
models such as BERT \cite{Devlin} and ERNIE \cite{Sun}, while~\citet{Puh} show
that jointly analyzing sentiment and ratings can provide more detailed insight.
Multilingual approaches and aspect-based methods further improve
interpretability by linking sentiment to specific components of the tourism
experience \cite{Chu2022,Ali}. More recently, zero-shot approaches have expanded the
feasibility of analyzing under-studied datasets with limited labeled data
\cite{Nawawi}.

This inconsistency shows up in regional literature as well. \citet{walgampaya} document rating–text incompatibility in hotel reviews in Anuradhapura, while~\citet{abeysinghe} extend this finding across five Sri Lankan cities and propose a self-learning approach to resolve it. However, both depend on lexicon-based methods and define the problem primarily as one requiring correction rather than explanation. In contrast, this study uses transformer-based sentiment analysis and interprets incongruence as a context-dependent NLP weak-label reliability issue.

In low-resource settings where a language does not have an adequate amount of tagged text sentiment data~\cite{de2026survey}, there have been attempts to derive the text sentiment using star ratings~\cite{jayawickrama2021seeking,jayawickrama2022facebook} or Facebook reactions~\cite{weeraprameshwara2022sinhala,weeraprameshwara2022sentiment}. 
However, empirical findings on the relationship between ratings and review text remain
mixed. \citet{Bigne} report general alignment between the two, but also
identify variation across contexts. \citet{George} show that ratings may
exceed text-based sentiment in destination-related reviews, while~\citet{Kwon} demonstrate that rating--text inconsistency varies by context and
influences perceived review usefulness. These findings suggest that ratings and
text do not always capture the same dimension of experience.

Reviewer characteristics also appear to matter. \citet{Chua} show that
reviewer expertise influences the relationship between ratings and textual
content, while related work links sentiment polarity and review depth to
perceived usefulness \cite{Chua,Ameur}. Taken together, these studies indicate that
ratings and text may encode different aspects of user experience and that
inconsistency may be partly shaped by reviewer-level behavior.

Overall, sentiment--rating incongruence remains insufficiently understood,
particularly in tourism attraction contexts and emerging destinations. Although
recent methods enable large-scale and fine-grained analysis \cite{Wen,Puh,Chu2022,Ali,Nawawi}, there
is still limited evidence on the structure of directional mismatch patterns and
their drivers in multi-venue, longitudinal datasets.
\section{Methodology}


\begin{figure*}[htbp]
\centering
\includegraphics[width=\linewidth]{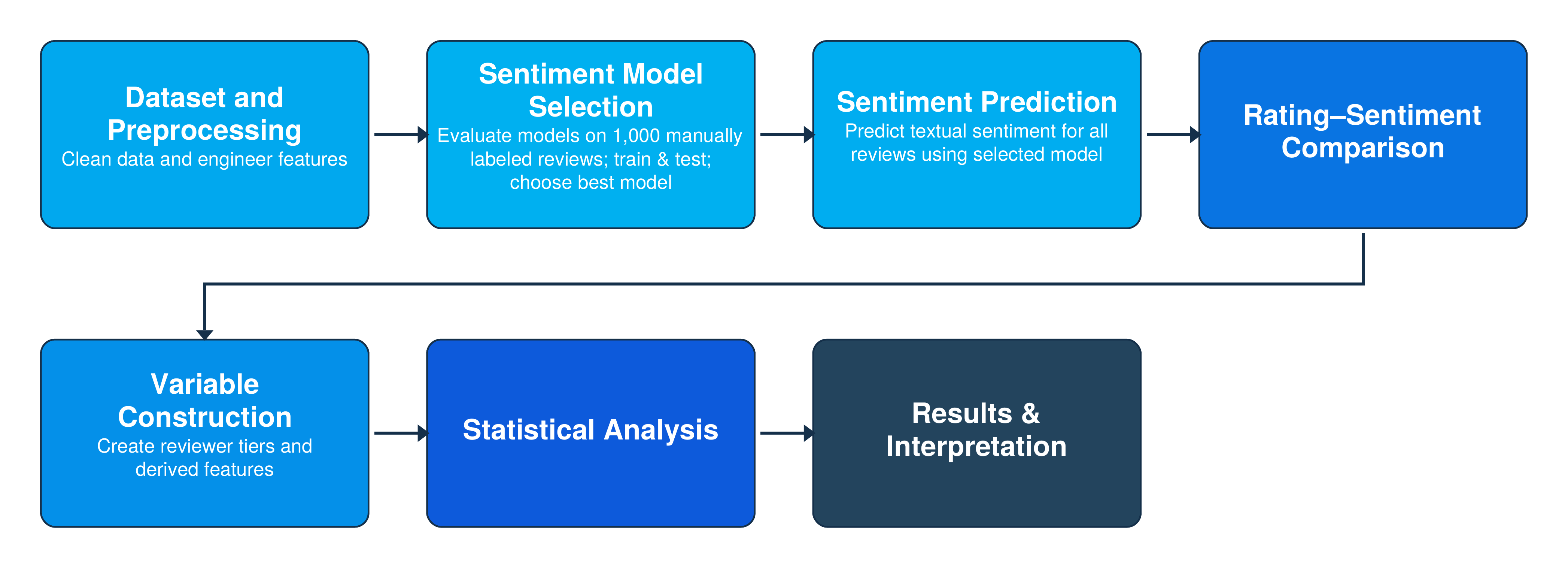}
\caption{Overview of the four-phase methodology.}
\label{fig:methodology_overview}
\end{figure*}

Fig.~\ref{fig:methodology_overview} summarizes the four-phase methodology adopted in this study.

\subsection{Dataset and Preprocessing}

The study used the ``Tourism and Travel Reviews: Sri Lankan
Destinations'' dataset from Mendeley Data~\cite{Sewwandi}, which contains 16,156 English reviews from 2010 to 2023 across 11 attraction types in Sri Lanka. Date fields were used
to create travel year and review delay, with negative delay values set to zero.
Raw location text was processed using rule-based parsing and manual mapping to
identify province and district. \textit{Review\_Length} was calculated as the
character count of the review text. Star ratings were grouped into three
classes:
\begin{itemize}
    \item Negative (1--2$\star$)
    \item Neutral (3$\star$)
    \item Positive (4--5$\star$)
\end{itemize}
This grouping follows common practice in sentiment
analysis~\cite{Pang,alaei} and makes the rating scale directly comparable with
the three-class sentiment output. Table~\ref{tab:dataset_attributes} lists the source columns retained for analysis.

\begin{table}[htbp]
\caption{Source Columns Used in Analysis}
\label{tab:dataset_attributes}
\centering
\resizebox{\columnwidth}{!}{%
\begin{tabular}{L{3.0cm}L{4.8cm}}
\toprule
\textbf{Feature} & \textbf{Description} \\
\midrule
Location            & Used to derive province and district \\
Location\_Type      & Type of attraction (museum, beach etc.) \\
User\_Contributions & Total reviews posted by the reviewer \\
Travel\_Date        & Date of visit; used to derive travel year \\
Published\_Date     & Date of review posting; used to derive review delay \\
Rating              & Star rating used to create Rating\_Class \\
Title               & Review title; combined with text for sentiment inference \\
Text                & Review body; used for sentiment and review length \\
\bottomrule
\end{tabular}%
}
\end{table}

\subsection{Sentiment Model Selection}

Model candidates were selected to span a representative range of adaptation strategies: a community fine-tuned checkpoint (gosorio/robertaSentimentFT) representing an off-the-shelf fine-tuned option, the pretrained cardiffnlp/twitter-roberta-base-sentiment model representing a strong general-purpose social-media sentiment baseline, and two in-house fine-tuned variants (Cardiff RoBERTa and RoBERTa-base) to test whether further domain-specific fine-tuning on the manually labeled subset improved performance. 

To derive textual sentiment independently from ratings, four transformer-based models were tested on a manually labeled set of 1,000 reviews. The dataset was split into 700 training and 300 testing instances, where the training portion was used to fine-tune selected models and the test set was used for comparative evaluation. Review titles and texts were combined as input, and model performance was assessed using Macro F1, accuracy, and weighted F1. Macro
F1 was included because sentiment classes may be imbalanced. As shown in Table~\ref{tab:sentiment_performance}, the pretrained \texttt{cardiffnlp/twitter-roberta-base-sentiment} model performed best overall and was selected to label the full dataset \cite{Devlin, Sun}. This model achieved a strong balance between classification performance and generalization, outperforming fine-tuned variants while avoiding potential overfitting given the limited size of the labeled dataset. This approach measured textual sentiment independently from star ratings, reducing circularity and enabling mismatch detection between two signals: rating-derived sentiment and NLP-derived textual sentiment.

\begin{table}[htbp]
\caption{Sentiment Model Performance (Test Set, $n = 300$)}
\label{tab:sentiment_performance}
\centering
\resizebox{\columnwidth}{!}{%
\begin{tabular}{L{4.2cm}cccc}
\toprule
\textbf{Model} & \textbf{Acc.} & \textbf{Mac. F1} & \textbf{Wt. F1} & \textbf{Decision} \\
\midrule
gosorio/robertaSentimentFT  & 0.730 & 0.281 & 0.616 & Rejected \\
cardiffnlp/twitter-roberta (pretrained) & 0.830 & 0.692 & 0.805 & \textbf{Selected} \\
Fine-tuned Cardiff RoBERTa  & 0.733 & 0.677 & 0.758 & Lower overall fit \\
Fine-tuned RoBERTa-base     & 0.777 & 0.698 & 0.780 & Higher Mac.\ F1 only \\
\bottomrule
\end{tabular}%
}
\end{table}

\subsection{Variable Construction}

After sentiment labeling, raw title and text were removed from further
analysis. \textit{Incongruent} was defined as a mismatch between
\textit{Sentiment} and \textit{Rating\_Class}, while \textit{Pattern} recorded
the six mismatch types. \textit{Reviewer\_Tier} grouped reviewers as follows: Novice (0--5), Casual (6--20), Active (21--100), Expert (101+).

Reviewer tiers were defined by analyzing the distribution of contributions. The data show strong positive skew ($\text{median} = 54$, $\text{max} = 9010$): most reviewers contribute $1$--$5$ reviews, while few exceed $100$. Based on this distribution, thresholds were set at $0$--$5$, $6$--$20$, $21$--$100$, and $101+$ to capture distinct engagement levels. Each tier corresponds to measurable differences in behavior. Reviewers with $0$--$5$ reviews exhibit minimal platform familiarity, whereas the $6$--$20$ range reflects casual engagement. The $21$--$100$ tier identifies active users with sustained participation, and $101+$ represents highly engaged expert reviewers. This tiering is further supported by rating behavior: the Conservative Rater pattern increases from $27.4\%$ among novice reviewers to $40.4\%$ among experts, indicating experience-dependent rating practices. These tiers therefore capture both contribution intensity and observable differences in rating–text alignment \cite{Chua}.

The constructed analytical variables used in this study comprise both target-defining and explanatory features. \textit{Sentiment} is represented as a three-class label derived from model output, while \textit{Rating\_Class} is a three-class label obtained by grouping star ratings (1--5); together, these variables define sentiment–rating incongruence, from which the binary variable \textit{Incongruent} (0/1) is derived as the target outcome. \textit{Pattern} is a six-category variable formed from the interaction between \textit{Sentiment} and \textit{Rating\_Class}, capturing distinct mismatch typologies, and \textit{Reviewer\_Tier} is a four-level ordinal variable based on grouped contribution levels, used as a predictor. Continuous predictors include \textit{log\_review\_length}, computed as the logarithm of review length in characters using $\log(1+x)$, and \textit{log\_review\_delay}, defined as the logarithm of the time gap between visit and posting using $\log(1+x)$. Temporal effects are modeled using \textit{Travel\_Year\_c}, a centered travel year variable, and its squared term \textit{Travel\_Year\_c\textsuperscript{2}}, which captures potential nonlinear time trends. Together, these variables capture key textual, behavioral, and temporal factors relevant to the analysis.

\begin{table}[htbp]
\caption{Modeling and Interpretation Framework}
\label{tab:model_framework}
\centering
\resizebox{\columnwidth}{!}{%
\begin{tabular}{L{1.6cm}L{2.5cm}L{1.4cm}L{2.8cm}}
\toprule
\textbf{Model} & \textbf{Purpose} & \textbf{Metric} & \textbf{Key Detail} \\
\midrule
1A: Logistic Reg.        & Linear baseline performance     & AUC-ROC       & Stratified 80/20 split; balanced classes; 5-fold CV \\
1B: Logit (statsmodels)  & Identifies independent drivers  & 95\% CI       & Unweighted inference model fit on full design matrix \\
2: Random Forest         & Captures complex relationships  & AUC-ROC       & GridSearchCV on training set; same held-out test set \\
SHAP (post hoc)          & Explains model predictions      & Mean $|\phi|$ & TreeExplainer on the best Random Forest \\
\bottomrule
\end{tabular}%
}
\end{table}

\subsection{Statistical Testing and Predictive Modeling}

After deriving textual sentiment through transformer-based NLP inference, statistical and machine learning models were used as secondary explanatory tools to examine factors associated with rating--text mismatch. \textit{Incongruent} was used as the binary outcome variable, while variables used to define it were excluded from the predictors to avoid data leakage. The final predictor set included venue type, province, reviewer tier, review length, review delay, and travel year terms, with low multicollinearity (max VIF~$= 3.663$).

Table~\ref{tab:model_framework} summarizes the modeling and interpretation framework used in this phase.
Chi-square and Mann--Whitney U tests were used for bivariate analysis, with Benjamini--Hochberg correction applied for multiple testing. Logistic regression and logit models were used to examine linear effects and adjusted odds ratios, while Random Forest was used to assess nonlinear relationships. SHAP was used only as a post hoc interpretation layer for the Random Forest model, supporting the explanation of NLP-derived incongruence rather than replacing the main sentiment analysis framework \cite{Puh}. Model performance was evaluated using AUC-ROC.

\section{Results}
\subsection{Prevalence and Six-Pattern Typology}

\begin{figure}[htbp]
\centering
\includegraphics[width=0.8\columnwidth]{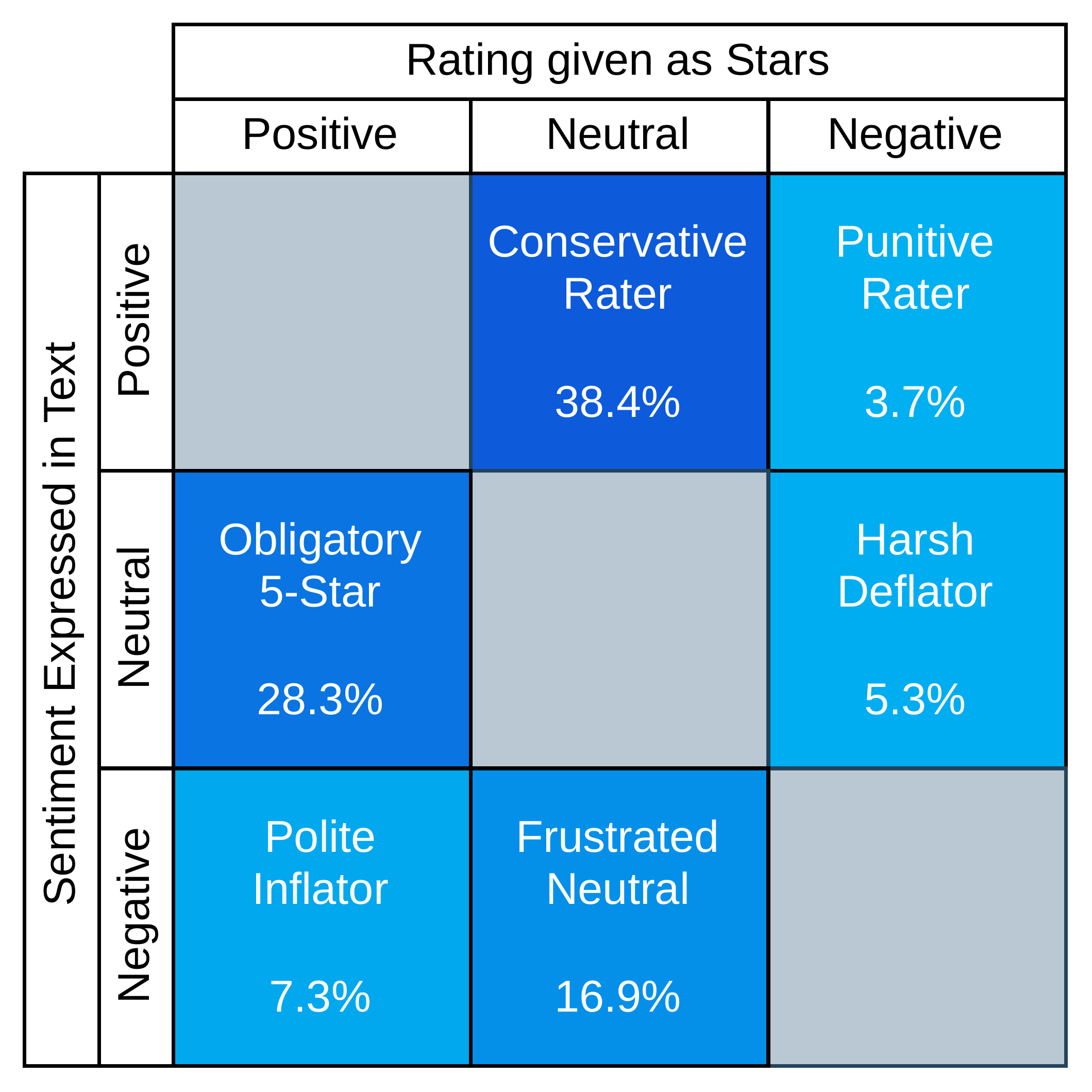}


\caption{Distribution of the six directional incongruence patterns.}
\label{fig:incongruence_patterns}
\end{figure}

Each incongruence type reflects a directional mismatch between rating and sentiment polarity. The pattern names, as shown in Fig~\ref{fig:incongruence_patterns}, are original to this study, derived from the behavioral characteristic each mismatch most plausibly reflects. 

Among the reviews analyzed, 3,005 were identified as
incongruent, giving an overall prevalence of 18.6\%, or roughly one in five
reviews. Fig.~\ref{fig:incongruence_patterns} further shows that incongruence follows a
six-pattern
typology. The two most common patterns, Conservative Rater (38.4\%) and
Obligatory 5-Star (28.3\%), together account for 66.7\% of all incongruent
reviews. This indicates that rating--text mismatch is directionally structured
rather than random. In addition, Frustrated Neutral and Polite Inflator account
for a further 24.2\% of incongruent cases, showing that negative sentiment is
often paired with non-negative ratings. This directional structure suggests that rating-derived labels introduce systematic rather than random noise into sentiment analysis tasks. The concentration of mismatches in a few recurring patterns also makes the typology useful for interpreting how numerical ratings and written sentiment diverge in review-mining datasets.

\subsection{Variation Across Venue Types}
\begin{figure}[htbp]
\centering
\includegraphics[width=\columnwidth]{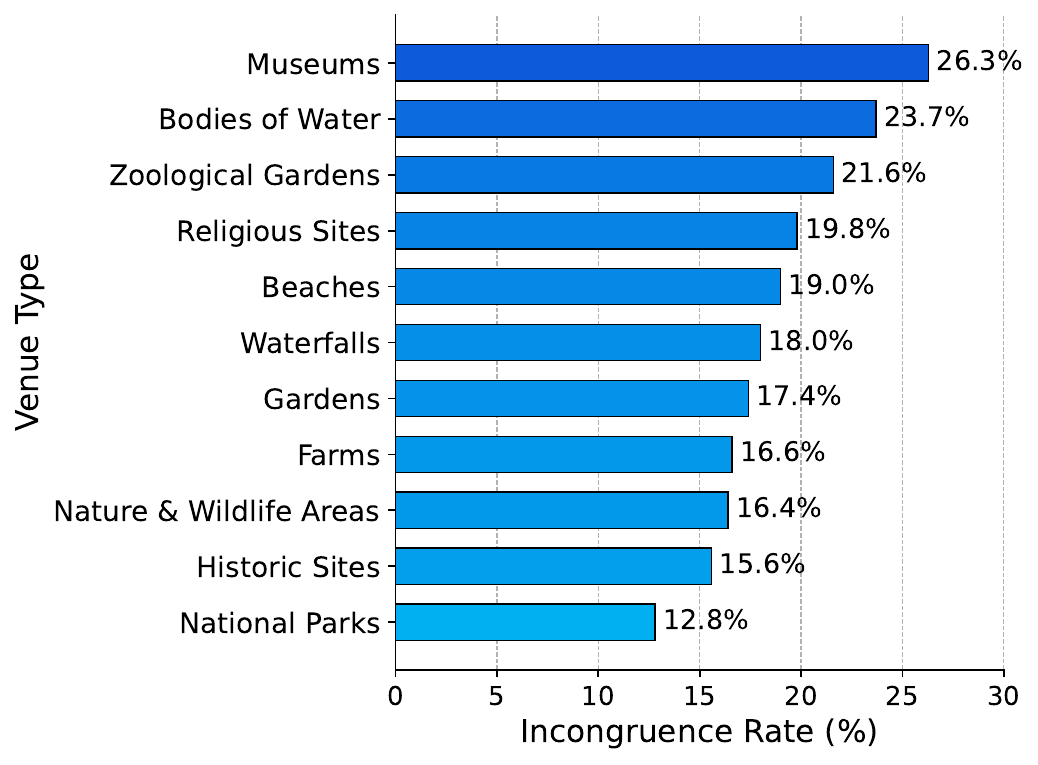}
\caption{Incongruence rate by venue type.}
\label{fig:venue_incongruence_rate}
\end{figure}

Incongruence rates were compared across 11 attraction categories to assess
contextual variation. The Chi-square test showed a statistically significant
association between venue type and incongruence, with a small but meaningful
effect size ($\chi^2(10) = 125.85$, $p < 0.001$; Cramer's V = 0.088). As shown
in Fig.~\ref{fig:venue_incongruence_rate}, National Parks had the lowest
incongruence rate (12.8\%), whereas Museums had the highest (26.3\%). Overall, variation across
venue types indicates that rating--text mismatch is not evenly distributed, but
differs by review context.

\subsection{Predictors of Incongruence}
Bivariate screening with Benjamini--Hochberg correction was used to identify
predictors associated with incongruence. As shown in
Table~\ref{tab:bivariate_screening}, reviewer tier, province, travel year, and
review length remained significant after correction,
while review delay was not significant ($q = 0.7503$). Expert reviewers were
1.97 times more likely than novices to produce incongruent reviews \cite{Chua}. In
addition, incongruent reviews were longer at the median than congruent reviews
(296 vs. 279 characters). These findings indicate that both reviewer
characteristics and review content are associated with sentiment--rating
mismatch, although multivariable modeling is needed to test their independent
effects.

\begin{figure}[htbp]
\centering
\includegraphics[width=0.9\columnwidth]{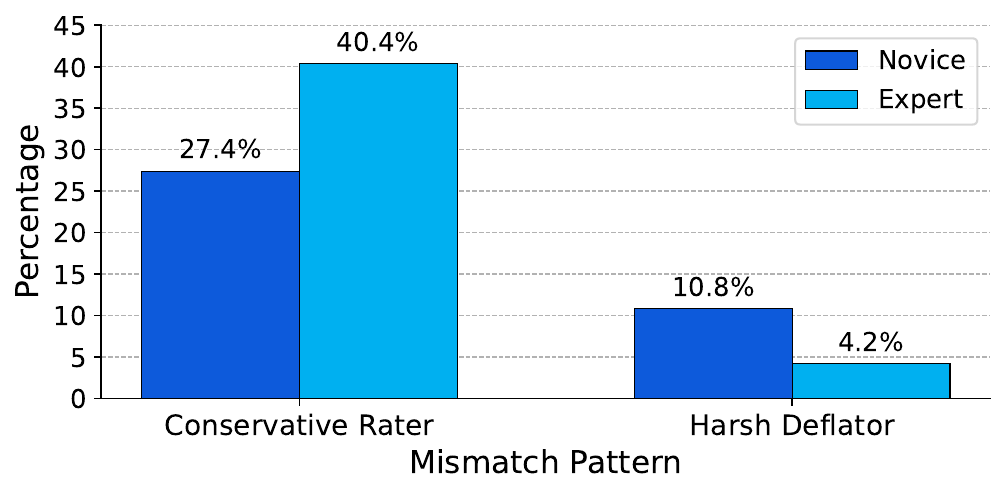}
\caption{Selected incongruence patterns by reviewer expertise, showing higher Conservative Rater prevalence and lower Harsh Deflator prevalence among expert reviewers.}
\label{fig:reviewer_expertise_patterns}
\end{figure}

Fig.~\ref{fig:reviewer_expertise_patterns} further illustrates the reviewer expertise effect for selected mismatch patterns, showing that Conservative Rater becomes more common among expert reviewers, while Harsh Deflator becomes less common.

\begin{table}[htbp]
\caption{Bivariate Predictor Screening with BH-FDR Correction}
\label{tab:bivariate_screening}
\centering
\resizebox{\columnwidth}{!}{%
\begin{tabular}{L{2.0cm}L{1.8cm}L{1.3cm}L{1.3cm}L{1.0cm}}
\toprule
Predictor & Test Statistic & Raw $p$ & $q$-value & Retained \\
\midrule
Reviewer Tier & $\chi^2(3)=85.99$ & $1.59\times10^{-18}$ & $7.94\times10^{-18}$ & Yes \\
Province & $\chi^2(7)=49.10$ & $2.17\times10^{-8}$ & $3.62\times10^{-8}$ & Yes \\
Travel Year & Mann--Whitney & $7.73\times10^{-11}$ & $1.93\times10^{-10}$ & Yes \\
Review Length & Mann--Whitney & 0.0011 & 0.0014 & Yes \\
Review Delay & Mann--Whitney & 0.7503 & 0.7503 & No \\
\bottomrule
\end{tabular}%
}
\end{table}

\subsection{Model-Based Analysis and Interpretation}
\text{1) Model 1A -- Logistic Regression:}
A class-balanced logistic regression model was used as a linear baseline. It
achieved a mean cross-validated AUC of $0.5890 \pm 0.0093$ and a test AUC of
0.5840, indicating modest but stable predictive performance. This suggests that
incongruence is only partly explained by the observed variables.

\text{2) Model 1B -- Explanatory Logit:}
Used to identify independent predictors of
incongruence. Based on 95\% confidence intervals, 19 predictors were
statistically significant. Venue type, reviewer expertise, review length, and
travel year showed important effects, while review delay had only a weak
negative association. Overall, the model shows that incongruence is shaped by
structural, behavioral, and temporal factors.

\text{3) Model 2 -- Random Forest (Nonlinear Structure Test):} Applied to capture nonlinear relationships and
interactions. It achieved a test AUC of 0.6095, which was slightly higher than
logistic regression. This indicates that nonlinear effects are present, although
their contribution is modest.

\text{4) SHAP Analysis:}
SHAP was used as a post hoc interpretation layer for the Random Forest model to identify influential features behind NLP-derived incongruence. The results were broadly consistent with the logit model, highlighting review length, reviewer expertise, travel year, review delay, and venue type as influential factors. SHAP therefore supports interpretation of nonlinear patterns rather than replacing the main sentiment analysis framework.

Temporal effects indicate a modest but consistent decline in sentiment–rating incongruence over time. The linear travel year term shows a significant negative association (Travel\_Year\_c: OR = 0.946, CI < 1), suggesting that more recent reviews are less likely to be incongruent. In contrast, the quadratic term (Travel\_Year\_c²: OR = 0.995) is not statistically significant, providing no evidence of nonlinear temporal effects and indicating an approximately linear trend. SHAP analysis further supports this finding, identifying Travel\_Year\_c as an influential predictor (mean |SHAP| = 0.0219) and confirming the overall direction of the effect. This indicates increasing alignment between textual sentiment and ratings over time.

\section{Discussion}
\subsection{Structured Incongruence}
The incongruence rate of 18.6\% shows that rating--text mismatch is systematic rather than random. Ratings capture overall judgments, while review text captures more specific details of the visitor experience. For sentiment analysis, a numerical score may compress a complex experience into a single label, while the written review can express mixed or context-dependent sentiment.

The six-pattern structure further shows that incongruence is directional rather than accidental. \textit{Conservative Rater} and \textit{Obligatory 5-Star} patterns dominate the mismatched cases, indicating that reviewers do not simply make random rating errors. For NLP pipelines, this means that star ratings do not function equally across contexts as weak sentiment labels. Using them without validation can introduce systematic label noise into sentiment analysis models. Ratings and textual sentiment therefore capture different dimensions of experience, and treating them as interchangeable signals creates modeling risk \cite{Bigne}. This supports the use of review text as an independent sentiment signal when building or evaluating review-mining systems. It also suggests that rating-based labels should be checked for domain-specific bias before being used for supervised sentiment classification.

\subsection{Conservative Rating Behavior}

The Conservative Rater pattern further reinforces structural incongruence. Positive textual sentiment paired with moderate ratings accounts for 38.4\% of incongruent reviews, showing that reviewers often temper numerical scores relative to their written sentiment. This pattern rises from 27.4\% among novice reviewers to 40.4\% among experts, suggesting that rating behavior becomes more calibrated with experience. Experienced reviewers may reserve high ratings for exceptional cases while still expressing positive textual sentiment. For NLP, this creates structured label noise when ratings are used as ground-truth sentiment labels, making reviewer expertise important for interpreting rating--text relationships.

\subsection{Location Type as a Structural Moderator of Label Reliability}
Location type affects rating–text reliability. Museums are over twice as likely to be incongruent compared with national parks (Adjusted OR = 2.386), with similar patterns for beaches, inland waterbodies, zoological gardens, and waterfalls. This suggests that some attraction types are harder to evaluate using a single numerical score. Museums and cultural sites often involve layered experiences, where visitors may describe positive exhibits, heritage value, or emotional significance while also mentioning issues such as crowding, pricing, accessibility, or facilities.

These findings indicate that star-rating reliability is context-dependent rather than uniform across all review types. Rating-derived labels may therefore introduce systematic bias when treated as equally reliable sentiment labels across different tourism contexts. For NLP, this supports the need for context-aware sentiment modeling that validates whether rating-derived sentiment and textual sentiment are aligned before using ratings as ground-truth labels. In practical NLP applications, this means that attraction categories may require different levels of label validation before ratings are reused as sentiment labels.

\subsection{Complementary Roles of Linear and Nonlinear Models}

The results show that linear and nonlinear models offer complementary insights into sentiment–rating incongruence rather than competing explanations. Logistic regression ($\mathrm{AUC} = 0.584$) provides a stable and interpretable baseline, identifying 19 significant predictors, while the Random Forest achieves a modest improvement ($\mathrm{AUC} = 0.6095$), confirming the presence of nonlinear and interaction effects. However, the limited performance gain suggests that these nonlinearities are not dominant. The overall predictive range ($\mathrm{AUC} \approx 0.58$--$0.61$) indicates that incongruence is structured but only partially observable, with substantial variation driven by latent behavioral and contextual factors. Importantly, this moderate predictive performance should not be interpreted as a weakness; rather, it reflects the inherent complexity and subjectivity of human judgment in review behavior, where not all influencing factors are directly measurable. SHAP analysis reinforces this interpretation by showing that variables such as reviewer expertise, review length, and venue type contribute in nonlinear and context-dependent ways, underscoring the need for richer features to further capture the phenomenon.

A further limitation concerns the reliability and suitability of the underlying sentiment model. Because the selected model (cardiffnlp/twitter-roberta-base-sentiment) is used in a pretrained, off-the-shelf form rather than fine-tuned on tourism-domain text, its predictions are shaped by its original Twitter-style training distribution; this raises a validity concern, since any systematic bias in how the model interprets sentiment may directly affect whether a given rating-text pair is classified as incongruent, independent of actual reviewer behavior. The selected model was originally trained on short-form, Twitter-style text, whereas tourism reviews are typically longer, more descriptive, and often express mixed or aspect-level sentiment within a single review; this domain mismatch may cause the model to misclassify nuanced or compound sentiment expressions, potentially inflating or deflating the measured incongruence rate. 
Compounding this concern, model evaluation itself relied on a manually labeled subset of only 1,000 reviews (700 training, 300 test), with model selection performed directly on the test set rather than a separate validation set. Given that the final claims are drawn over the full corpus of 16,156 reviews, this relatively small and unvalidated evaluation basis means that the selected model's generalization cannot be fully verified. 

Furthermore, the reported incongruence patterns rely on a single sentiment model and a fixed modeling framework (logistic regression, Random Forest, and SHAP); robustness against alternative baselines (e.g., VADER, TextBlob, TF-IDF-based classifiers, or domain-specific review sentiment models) and alternative predictive models (e.g., XGBoost, SVM) was not assessed. Future work should incorporate a larger manually annotated sample with a dedicated validation split, explore domain adaptation or fine-tuning on review-specific corpora, and benchmark against these alternative baselines and modeling approaches to establish whether the observed patterns are robust or specific to the chosen pipeline.

Additionally, because incongruence is defined purely as a mismatch between rating and model-predicted sentiment, some flagged cases may reflect sentiment model error rather than genuine reviewer behavior. No manual validation of the incongruent subset was performed in this study; a manual audit of a sample of incongruent reviews would help confirm that the identified behavioral patterns reflect actual reviewer tendencies rather than residual classification noise.

\subsection{Review Delay and Reviewer Expertise Effects }

Review delay plays limited role in incongruence, showing no significance in bivariate analysis and only a weak negative association (OR = 0.960). In contrast, reviewer expertise is more influential, with experts exhibiting more conservative and fewer harsh rating behaviors, indicating that incongruence is driven more by reviewer behavior than timing.

\FloatBarrier

\section{Conclusion}
Sentiment–Rating incongruence in tourism reviews is systematic rather than random. Using transformer-based sentiment inference, this study showed that 18.6\% of reviews contain rating–text mismatch forming six directional patterns. The findings show that reviewer expertise, review length, venue type, and temporal factors influence rating–text divergence. More importantly, star ratings and textual sentiment are not interchangeable signals \cite{Bigne}. This distinction is especially important for datasets where ratings are used automatically as training labels.

For NLP research, the main implication is that star ratings should not be treated as ground-truth sentiment labels without validation. Rating-derived sentiment labels may introduce systematic label noise when the written review expresses a different sentiment polarity from the assigned score. This study therefore supports the need for context-aware sentiment analysis methods that evaluate weak-label reliability before model training or evaluation.
This puts the premise and validity of some prior low-resource sentiment analysis work~\cite{jayawickrama2021seeking,jayawickrama2022facebook} into question. 

{\footnotesize
\bibliographystyle{IEEEtranN}
\bibliography{references}
}

\end{document}